\documentclass[runningheads]{llncs}
\usepackage[T1]{fontenc}
\usepackage{graphicx}
\usepackage{booktabs}
\usepackage{siunitx}
\usepackage{adjustbox}
\usepackage{tabularx}
\usepackage{pifont}
\usepackage{array}
\usepackage{xcolor}
\usepackage{url}
\usepackage{amsfonts}

\usepackage{multirow}
\usepackage{float}

\usepackage{tikz}
\usetikzlibrary{
    positioning,
    arrows.meta,
    shapes.geometric,
    calc
}
\usepackage{float}
\usepackage{tcolorbox}
\usepackage{listings}
\begin{document}
\title{Runtime Uncertainty Monitoring for LLM-Based Multi-Agent Systems Using Bayesian Networks}
\titlerunning{Uncertainty Monitoring for LLM Agents}
%
\author{Bart Custers\inst{1}\orcidID{0009-0005-7507-4690} \and
Koorosh Aslansefat\inst{1,2}\orcidID{0000-0001-9318-8177}}
\authorrunning{B. Custers et al.}
%
\institute{University of Hull, Hull HU6 7RX, UK
\email{bart\_custers@hotmail.com}\\
\and
\email{k.aslansefat@hull.ac.uk}}
\maketitle              
\begin{abstract}
This paper investigates how multi-agent systems (MAS)-based on large language models (LLMs) can support actuarial risk modelling, with a particular focus on uncertainty quantification. Actuarial workflows represent a high-stakes decision-support setting where unreliable outputs may lead to incorrect risk assessment, unfair pricing, and regulatory non-compliance. To address uncertainty introduced by the probabilistic nature of LLMs and dependencies between agents, a multi-agent framework is proposed in which specialised agents perform data preparation, modelling, review, and explanation tasks under a central hub. The main contribution is a novel approach to uncertainty propagation using token-level log-probabilities and a Bayesian Network. Importantly, log probabilities are not treated as direct probabilities of correctness or task success. Instead, length-normalised log-probability summaries are transformed into calibrated task-level confidence estimates before incorporation into the Bayesian Network. Results show that the framework reproduces baseline actuarial performance while providing additional insight into workflow stability and runtime uncertainty propagation.

\keywords{Multi-Agent Systems  \and Uncertainty Quantification \and Runtime Monitoring} \and Actuarial Modelling
\end{abstract}

\section{Introduction}
Over the past decades, artificial intelligence (AI) has become increasingly important in the insurance industry, particularly for predictive modelling and automation \cite{Owens2022}. Accurate risk estimation is central to insurance, as insurers rely on predictions of future claims to determine premiums and ensure both profitability and solvency. Risk modelling aims to capture differences in risk across policyholders, as these differences directly influence pricing decisions. For example, characteristics such as age or driving experience can significantly affect claim frequency, making accurate risk differentiation essential. Despite ongoing technological progress, actuarial risk modelling remains a resource-intensive process that requires substantial manual effort. 

Recent advances in large language models (LLMs) have enabled the development of AI agents capable of performing complex reasoning tasks \cite{Wang2024,Sapkota2025}. These agents can potentially automate actuarial workflows, which typically involve data preparation, model development, validation, and interpretation. However, the use of LLM-based agents in actuarial risk modelling remains largely unexplored. A key challenge in adopting such systems is their lack of transparency. AI models are often perceived as “black boxes,” and LLM-based agents introduce additional concerns, including hallucinations and variability across runs \cite{Wang2024}. For actuaries, this is problematic, as transparency and explainability are essential in a highly regulated environment \cite{InternationalActuarialAssociation2024}.  

\textcolor{black}{The safety implications of LLM-based multi-agent systems are particularly important in insurance risk modelling, where automated decisions may directly affect pricing, customer fairness, and financial accessibility. Errors introduced by a single agent could propagate to subsequent workflow stages, potentially leading to incorrect or biased outcomes that may be difficult to detect. Existing approaches do not sufficiently address how errors or uncertainty arise and propagate in multi-step, agent-based workflows.} In addition to incorporating explainability techniques such as concept-based explanations and fairness assessments, this work therefore introduces a novel uncertainty modelling approach. \textcolor{black}{We propose a runtime safety-monitoring framework for LLM-based multi-agent systems in high-stakes insurance workflows.} The main contribution lies in combining token-level log probabilities from LLM outputs with a Bayesian Network that captures dependencies between workflow stages. As a result, the framework provides a structured representation of uncertainty, and a safety mechanism that may help identify unsafe workflows states.

\section{Background}
\subsection{Multi-Agent Systems}
Multi-agent systems (MAS) extend the concept of single LLM-based agents by enabling multiple agents to collaborate on shared tasks. This collaborative setup is particularly useful for complex problem-solving, where tasks can be decomposed into smaller, specialized components, which improves overall system performance \cite{Yang2024}. In addition, MAS can enhance robustness, since failures in one component may be mitigated by other agents within the system \cite{Yang2024,Tran2025}.  Frameworks such as AutoGen \cite{Wu2023} and AgentVerse \cite{Chen2023}, for example, show how various agents are able to collaborate on complex tasks. Applications of MAS in the insurance domain, particularly for actuarial tasks, remain relatively limited. Existing studies primarily focus on the use of LLMs for specific tasks, such as processing unstructured data or supporting actuarial analysis \cite{Hatzesberger2025}. Despite all advances, MAS face several challenges. LLM-based agents are not inherently trained to collaborate, which may result in unpredictable behaviour or inconsistent reasoning \cite{Tran2025}. Furthermore, issues such as hallucination and sensitivity to prompts persist in multi-agent settings.

Evaluating LLM-based MAS is challenging due to their dynamic and probabilistic nature. Traditional evaluation metrics, which focus on task-specific performance, are often insufficient for capturing the broader behaviour of agent systems. Recent research therefore emphasizes holistic evaluation approaches that consider multiple aspects of system performance \cite{Mohammadi2025,Yehudai2025}.
One such framework distinguishes between evaluation objectives and evaluation processes \cite{Mohammadi2025}. Evaluation objectives define what should be assessed, including dimensions such as agent behaviour, capabilities, reliability, and safety or alignment. Despite these developments, several limitations remain. Many evaluation approaches still focus on narrow aspects such as task accuracy or tool usage, while neglecting system-level properties like robustness and interaction dynamics \cite{Mohammadi2025}. 

\subsection{Uncertainty quantification}
A key consideration in deploying MAS is the handling of uncertainty. In autoregressive language models, uncertainty arises from the probability distribution over possible next tokens during text generation \cite{Malinin2020}. Token-level log probabilities provide a model-internal confidence-related signal, but not a direct probability of correctness. The use of log probabilities ("logprobs") has been demonstrated by \cite{Chauvin2025}, where the authors propose Logprobs to monitor LLM updates. Furthermore, \cite{Xu2025} use logprobs to calculate uncertainty and reduce LLM hallucinations.
However, such uncertainty estimates do not capture how uncertainty accumulates across the workflow. In MAS, decisions are interdependent, meaning that the output of one agent influences subsequent steps. This leads to a distinction between intrinsic uncertainty, which reflects uncertainty at a single step, and extrinsic uncertainty, which captures uncertainty propagated from earlier decisions \cite{Duan2025}. 
Sampling-based approaches, such as Monte Carlo methods, have been proposed to approximate uncertainty propagation by generating multiple reasoning trajectories \cite{Malinin2020}. Although these methods can provide insights, they are computationally intensive and may be difficult to interpret. An alternative approach is to represent dependencies explicitly using a Bayesian Network, where nodes correspond to agent decisions and edges represent their relationships \cite{Hu2024,Nafar2025,donaldson2026bayesian}. This allows uncertainty to be modeled in a structured and interpretable way.

\section{Multi-agent system design}

\begin{figure}[htbp!] 
\centering    
\includegraphics[width=0.9\textwidth]{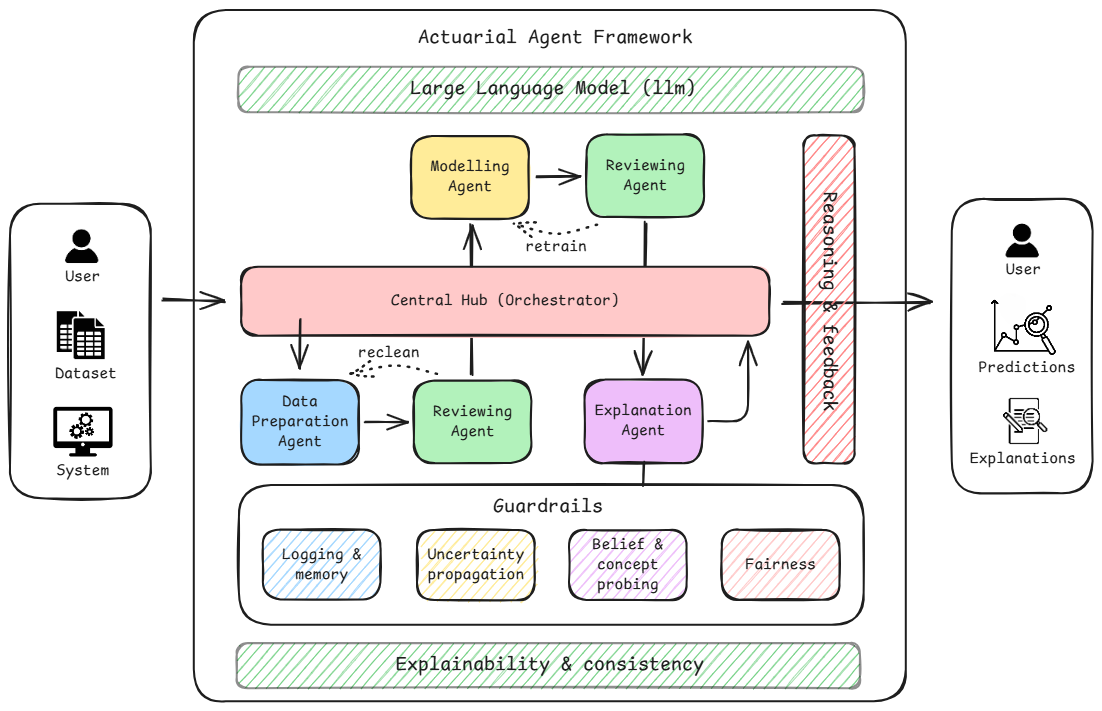}
\caption{Sketch of the application's architecture}
\label{fig:3.1}
\end{figure}

The framework, shown in Fig.~\ref{fig:3.1}, follows a centralised star topology \cite{Yang2024}. A Central Hub coordinates four specialised agents through message passing, avoiding direct agent-to-agent communication and enabling controlled workflow oversight. The Data Preparation Agent performs data ingestion, cleaning, and encoding. The Modelling Agent develops predictive models. The Reviewing Agent evaluates data preparation and modelling outputs. The Explanation Agent supports interpretability and guardrails. Each agent records outputs and metadata, while shared memory stores information from previous workflow stages.

The Central Hub routes tasks based on agent feedback. If the Reviewing Agent detects issues, the workflow can be revised or terminated. To reduce prompt complexity, each agent task is divided into layers such as planning, code generation, evaluation, and explanation. The Reviewing Agent checks performance, consistency, and modelling decisions. The Explanation Agent assesses internal beliefs, concept alignment, and fairness. Concept alignment is measured using Testing with Concept Activation Vectors \cite{Kim2017}, while fairness is assessed through group-level comparisons between predicted and observed outcomes for sensitive attributes such as age and population density.

The framework supports three LLM backends, all relatively small in size, as the aim was to invoke backends that can be quickly run on local machines. The three models, Llama 2 7B, and Llama 3.1 8B \cite{MetaAI2024}, and Qwen2.5 7B \cite{Qwen}, can all be characterized as decoder-only transformer models and are all available as open source model. 

\section{Uncertainty quantification}
\subsection{Log probabilities and uncertainty estimation}

Autoregressive language models generate outputs token-by-token from a predictive distribution conditioned on the input prompt. Given an input $x$ and model parameters $\theta$, the predictive distribution over outputs $y$ can be written as $p_{\theta}(y \mid x)$, and for a generated sequence of outputs $y=(z_1,\ldots,z_L)$ with length $L$, the probability can be decomposed into conditional token probabilities:

\begin{equation}
p_{\theta}(y|x)=\prod_{t=1}^{L}p_{\theta}(z_t|z_{<t},x)
\end{equation}

\noindent Taking the logarithm yields the log probability of the sequence:

\begin{equation}
\log p_{\theta}(y|x)=\sum_{t=1}^{L}\log p_{\theta}(z_t|z_{<t},x)
\end{equation}

\noindent Log probabilities ('logprobs') therefore provide a systematic approach to quantify how likely a particular outcome is according to the model, and they are directly derived from the predictive distribution. Because the text length of an input prompt can influence the magnitude of the log probability, it is common to apply length normalization to ensure fair comparison between responses of different lengths \cite{Duan2025}. Hence, the normalized log probability \(\bar{\ell}(y)\) is computed as:

\begin{equation}
\bar{\ell}(y)=\frac{1}{L}\sum_{t=1}^{L}\log p_{\theta}(z_t|z_{<t},x)
\end{equation}

Although log probabilities provide useful confidence-related signals, they do not directly represent the probability that an agent successfully completed its task, since LLMs may assign high likelihood to fluent but incorrect outputs. Therefore, the proposed framework treats log-probability summaries as uncertainty features rather than direct success probabilities. For agent $i$, the length-normalised log-probability summary is computed as
\begin{equation}
s_i = \frac{1}{L_i}\sum_{t=1}^{L_i}\log p_\theta(z_t|z_{<t},x),
\end{equation}
which is transformed into a bounded confidence feature $q_i=\exp(s_i)$. The probability of successful task completion is then estimated as
\begin{equation}
P(X_i = success \mid q_i) = f_{\mathrm{cal}}(q_i)
\end{equation}
where $f_{\text{cal}}$ denotes a calibration function estimated using validation runs. The calibrated value is then incorporated into the Bayesian Network as probabilistic evidence.

\subsection{Bayesian Networks}
\label{sec:5.4}
\begin{figure}[htbp!] 
\centering    
\includegraphics[width=1\textwidth]{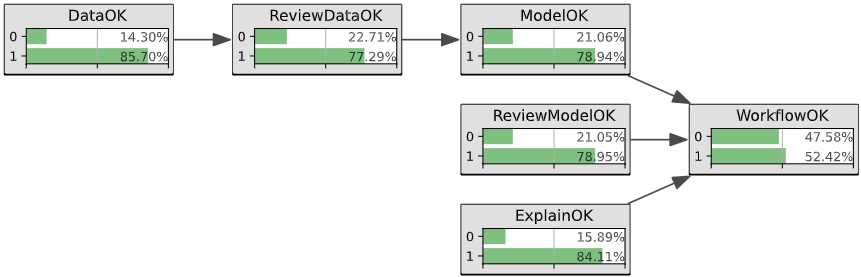}
\caption{Bayesian Network representation of the actuarial workflow, including posteriors for each phase success (1) or failure (0).}
\label{fig:4.1}
\end{figure}

In this MAS approach, uncertainty propagation is represented using a Bayesian Network that models the dependencies between agents. Each stage of the workflow is modelled as a node, and dependencies between stages are captured through directed edges, as shown in Fig. \ref{fig:4.1}. Token-level log probabilities are used to quantify uncertainty at each node, which is then propagated through the network to estimate overall system uncertainty. Bayesian Networks (BNs) are probabilistic graphical models that represent the joint probability distribution of a set of random variables using a directed acyclic graph \cite{Pearl1988}. Each node in the graph corresponds to a random variable, and directed edges represent conditional dependencies between variables. Let $X=\{X_1,...,X_n\}$ that denotes workflow tasks. For example, \(X_1\) may represent successful data preparation, \(X_2\) the successful review of the data, and \(X_3\) the successful training of a model.
The defining property of a Bayesian Network is that the joint distribution factorizes according to the structure of the graph:

\begin{equation}
P(X_1,...,X_n)=\prod_{i=1}^{n}P(X_i|\mathrm{Pa}(X_i))
\end{equation}

\noindent where \( \mathrm{Pa}(X_i) \) denotes the set of parent nodes of \(X_i\) \cite{Pearl1988}. Each conditional probability term is specified through a conditional probability table (CPT). These tables quantify how the probability of a node depends on its parents. Hence, Bayesian Networks are suitable frameworks to visualize how conditional dependence and uncertainty accumulate through a network.
In the context of the agent workflow, each node $X_i$ in the Bayesian Network represents the successful execution of a task performed by an agent. The confidence score obtained from the log probabilities of the agent’s response is used to parametrize the prior probability of that node, where $y_i$ denotes the output produced by agent $i$ is $P(X_i = \text{success}) = c(y_i)$.

\begin{figure}[htbp!] 
\centering    
\includegraphics[width=1\textwidth]{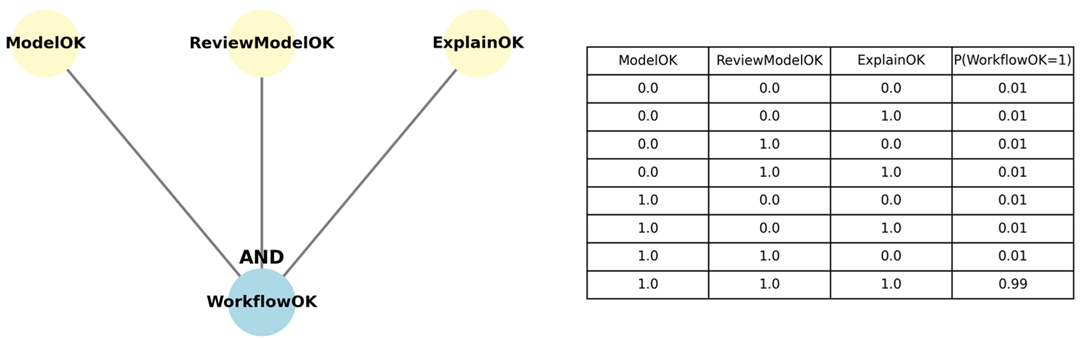}
\caption{CPT example of WorkflowOK, depending on three parent nodes. The “AND” gate label indicates that WorkflowOK succeeds only if all parent nodes pass.}
\label{fig:4.2}
\end{figure}

\noindent The Bayesian Network used in this study represents an abstracted execution trace of the workflow rather than the full control logic of the agentic system. We do not model temporal ordering beyond direct parent-child dependencies, nor do we explicitly represent task priority, repeated execution, feedback loops, or iterative revision cycles. If an agent requests a revision, the final accepted output of that stage is used as the evidence for the corresponding node. This assumption keeps the graph acyclic and interpretable, which is appropriate for the present proof-of-concept study. Extending the model to dynamic Bayesian Networks or influence diagrams is left for future work.
\noindent These values form the initial probability assignments for nodes that do not have parent variables. For nodes that depend on preceding steps in the workflow, the confidence score contributes to the conditional probability tables. As a CPT example, Fig. \ref{fig:4.2} shows that the relationship between WorkflowOK and its three parent nodes can be characterized as an "AND" gate, implying that WorkflowOK can only be true if all parent nodes are true. The CPT on the right shows the probability of WorkflowOK being true (1) for each combination of parent outcomes. When all parent nodes are true, the probability of WorkflowOK being true is equal to its prior (0.99), or posterior when evidence is gathered during the workflow process. 

Along the workflow run, new evidence is included in the network, and posterior probabilities are updated. The resulting network provides a dependency graph of uncertainty propagation within the workflow, as visualized in Fig. \ref{fig:4.1}. An important advantage of a Bayesian Network is its interpretability. The graphical structure makes it possible to identify which components of the workflow contribute most to overall uncertainty. If a particular node shows high uncertainty or strongly influences downstream tasks, it could be flagged as a potential problem in the workflow.

\subsection{Inference and uncertainty quantification}
Once the Bayesian Network structure and conditional probability tables are defined, probabilistic inference can be performed to estimate posterior probabilities across the workflow. The objective of this inference is to determine how uncertainty in individual agent outputs affects the reliability of the final workflow outcome. The Bayesian Network for the actuarial workflow stores prior probabilities, CPTs, and holds functions to update the network and for inference. Throughout the workflow, the agents log the uncertainty measurements in the metadata, which is then extracted by the Central Hub that updates the Bayesian Network accordingly. At the end of the workflow, this results in a Bayesian Network graph with conditional probabilities on workflow subtasks. 

\begin{table}[htbp]
\centering
\caption{Scenario-based probability of successful workflow execution.}
\label{tab:workflow_scenarios}

\begin{tabularx}{\textwidth}{p{1.6cm} X >{\centering\arraybackslash}p{3cm}}
\hline
\textbf{Scenario} & 
\textbf{Description} & 
\textbf{P(WorkflowOK)} \\
\hline

A
&
Safe scenario: low temperature (0.2) for all agents
& 87.1\%
\\
\hline

B
&
Mixed scenario: low temperature (0.2) for data prep and modelling agent, medium temperature (0.7) for reviewing and explanation agent
& 69.0\%
\\
\hline

C
&
Moderate scenario: medium temperature (0.7) for all agents (default)
& 59.2\%
\\
\hline

D
&
Risky scenario: high temperature (1.2) for one agent 
& 42.8\%
\\
\hline

\end{tabularx}
\end{table}

Fig. \ref{fig:4.1} shows an example of the Bayesian Network graph produced by the actuarial workflow. It visualizes for each step or agent in the workflow the inferred average and standard deviation of the logprobs, based on the underlying dependencies. Here, most steps in the workflow show a certainty around 80\%. However, the Bayesian Network also shows the propagation of uncertainty, as combining multiple agent outputs to finish the workflow results in a final certainty of 55\% in this presented example, which is significantly less compared to the individual agent certainties. To further test uncertainty propagation, the workflow can be tested with different settings, such as the structure (e.g. sequential), or LLM temperature. The scenarios for tests with different temperature settings for the LLM backends are shown in Table \ref{tab:workflow_scenarios}. By testing with different LLM temperatures, the amount of randomness in the agent's responses is varied, which has a direct effect on uncertainty. The scenarios in Table \ref{tab:workflow_scenarios} illustrate how increased randomness and uncertainty propagate through the network and reduce the probability that the overall workflow succeeds (1). In a conservative scenario (A), all agents operate with a low temperature (0.2). As expected, the probabilities of success are high across the network and the final WorkflowOK node shows the highest reliability. Scenario B introduces a mixed configuration: low temperature for the Data Preparation and Modelling agents, but medium temperature for Reviewing and Explanation agents. This increases variability in some tasks and leads to a decline in the probability that the workflow succeeds. Scenario C applies a moderate temperature (0.7) to all agents, thereby representing the default settings of the LLM backends. Finally, Scenario D maintains moderate temperatures but increases the temperature of the Explanation agent (1.2), showing how higher uncertainty in a single component can significantly reduce the reliability of the overall workflow. Hence, this approach provides useful insights into workflow certainty and possible weak elements in the workflow.  

\section{Evaluation}
The evaluation of the MAS focuses on four key dimensions: agent behaviour, consistency, agent capabilities (including error detection and adaptability), and safety. The analysis is based on repeated executions of the workflow using three LLM backends. Each backend was tested on the original dataset (10 runs) and on 20 systematically perturbed datasets designed to introduce controlled errors. Results are summarized in Table \ref{tab:combined_metrics_uncertainty}. In terms of general performance and agent behaviour, the MAS achieved a high task success rate, with at least 80\% of runs completed successfully across all backends. The predictive performance, measured using RMSE, was comparable to a baseline model, indicating that the MAS reliably reproduced standard actuarial results. While the Llama models performed similarly to the baseline, the Qwen backend achieved slightly better average performance. However, workflow certainty, derived from a Bayesian Network, varied significantly across backends, suggesting differences in how confidently each model evaluated its own outputs. Additionally, agent decisions differed across backends: Llama 2 tended to approve workflows more easily, while Llama 3.1 and Qwen more frequently included critical notes, highlighting variability in agent judgement despite identical prompts.





\begin{table}[ht]
\centering
\caption{Workflow performance metrics, perturbed test metrics, and agent-level uncertainty comparison of LLM backends}
\label{tab:combined_metrics_uncertainty}
\begin{adjustbox}{max width=\textwidth}
\begin{tabular}{l
                S[table-format=1.4]
                S[table-format=1.4]
                S[table-format=1.4]}
\toprule
\textbf{Metric} & 
\textbf{Llama 2 7B} & 
\textbf{Llama 3.1 8B} & 
\textbf{Qwen2.5 7B} \\
\midrule

\multicolumn{4}{l}{\textit{Main workflow metrics}} \\
Task Success Rate              & 0.8000 & 0.9000 & 0.8000 \\
Baseline model RMSE            & 0.8853 & 0.8853 & 0.8853 \\
Mean RMSE                      & 0.8856 & 0.8839 & 0.8671 \\
Mean workflow uncertainty      & 0.7345 & 0.5506 & 0.6012 \\
\midrule

\multicolumn{4}{l}{\textit{Test perturbed dataset metrics}} \\
Error Detection Rate           & 0.4500 & 0.6500 & 0.9000 \\
Mean RMSE                      & 1.0655 & 1.0728 & 1.0157 \\
Mean workflow uncertainty      & 0.7342 & 0.5573 & 0.5833 \\
\midrule

\multicolumn{4}{l}{\textit{Agent-level uncertainty scores}} \\
Data Preparation Agent         & 0.9226 & 0.8604 & 0.8648 \\
Reviewing Agent                & 0.9086 & 0.8107 & 0.8710 \\
Modelling Agent                & 0.9097 & 0.8292 & 0.8410 \\
Explanation Agent              & 0.8819 & 0.8429 & 0.8052 \\
Overall Workflow               & 0.7345 & 0.5506 & 0.6012 \\
\bottomrule
\end{tabular}
\end{adjustbox}
\end{table}


A key focus of the evaluation was error detection and adaptability under perturbed datasets. Here, substantial differences between backends emerged. The Qwen model achieved a high error detection rate (90\%), significantly outperforming Llama 2 (45\%) and Llama 3.1 (65\%). It also showed better predictive performance under perturbations and demonstrated a stronger tendency to adapt by retraining models or changing strategies. In contrast, the Llama models often failed to detect severe issues and rarely adjusted their approach. Further analysis showed that the MAS was particularly sensitive to distributional changes (population shifts), while other perturbations, such as missing values or sparsity, were less reliably detected. This highlights specific weaknesses in the validation process and suggests areas for improvement. Regarding safety, the Explanation Agent acts as a guardrail by evaluating belief consistency, interpretability (via TCAV), and fairness. The uncertainty tracking through the Bayesian Network provided useful insights. Table \ref{tab:combined_metrics_uncertainty}, the third part shows the average uncertainty scores per agent, and the combined uncertainty over the whole workflow. These results highlight how, even with high individual uncertainty scores for the agents, the propagation through the workflow can significantly reduce the score.


\begin{table}[ht]
\centering
\caption{Overview of test datasets, severity indicators and uncertainty scores}
\label{tab:dataset_perturbations}
\begin{adjustbox}{max width=\textwidth}
\begin{tabular}{ll
                S[table-format=1.4]
                S[table-format=1.4]
                S[table-format=1.4]
                S[table-format=1.4]
                S[table-format=1.4]
                S[table-format=1.4]
                S[table-format=1.4]
                S[table-format=1.4]}
\toprule
\textbf{ID} &
\textbf{Description} &
\textbf{Sparsity} &
\textbf{Row} &
\textbf{Column} &
\textbf{Missing} &
\textbf{Target} &
\textbf{Population} &
\textbf{Correlation} &
\textbf{Uncertainty} \\
&
&
&
\textbf{Change} &
\textbf{Change} &
\textbf{Rate} &
\textbf{Shift} &
\textbf{Shift} &
\textbf{Shift} &
\textbf{} \\
\midrule
T1  & Missing values - low         & 0.0946 & na      & na      & 0.0083 & na      & 0.0000 & 0.0016 & 0.5874 \\
T2  & Missing values - high         & 0.1279 & na      & na      & 0.0417 & na      & 0.0000 & 0.0035 & 0.5737 \\
T3  & Missing rows - low           & 0.0862 & 0.0295  & na      & na     & 0.0001  & 0.0000 & 0.0017 & 0.5734 \\
T4  & Missing rows - high           & 0.0862 & 0.1475  & na      & na     & 0.0008  & 0.0000 & 0.0047 & 0.5858 \\
T5  & Missing column - low         & 0.0863 & na      & 0.0833  & na     & na      & na     & na     & {0.6033}     \\
T6  & Missing column - high        & 0.0950 & na      & 0.1667  & na     & na      & na     & na     & {0.5737}     \\
T7  & Data increase - low          & 0.0863 & 0.0500  & na      & na     & 0.0001  & 0.0000 & 0.0018 & 0.5558 \\
T8  & Data increase - high          & 0.0862 & 0.2500  & na      & na     & 0.0004  & 0.0000 & 0.0035 & 0.5536 \\
T9  & Extra column - var1          & 0.0796 & na      & 0.0833  & na     & na      & na     & na     & {0.5648}     \\
T10 & Extra column - var2          & 0.0796 & na      & 0.0833  & na     & na      & na     & na     & {0.5837}     \\
T11  & Inject feature noise - var1   & 0.0862 & na      & na      & na     & na      & 0.0022 & 0.0071 & 0.5963 \\
T12 & Inject feature noise - var2  & 0.0862 & na      & na      & na     & na      & 0.0006 & 0.0066 & 0.5843 \\
T13 & Inject label noise - low     & 0.0823 & na      & na      & na     & 0.2093  & na     & 0.0574 & 0.6078 \\
T14 & Inject label noise - high    & 0.0705 & na      & na      & na     & 0.8303  & na     & 0.1074 & 0.6065 \\
T15 & Distribution shift - var1     & 0.0853 & 0.1653  & na      & na     & 0.0027  & 0.0130 & 0.0303 & 0.5785 \\
T16 & Distribution shift - var2    & 0.0861 & 0.0287  & na      & na     & 0.0026  & 0.0033 & 0.0584 & 0.5607 \\
T17 & Fairness test - low          & 0.0818 & na      & na      & na     & 0.2389  & na     & 0.5973 & 0.5897 \\
T18 & Fairness test - high         & 0.0740 & na      & na      & na     & 0.6504  & na     & 0.3707 & 0.6139 \\
T19 & Counterfactual test 1   & 0.0862 & na      & na      & na     & na      & na     & 0.7575 & 0.5727 \\
T20 & Counterfactual test 2   & 0.0862 & na      & na      & na     & na      & na     & 0.1612 & 0.5998 \\
\bottomrule
\end{tabular}
\end{adjustbox}
\end{table}

Each backend was also tested on 20 perturbed datasets to assess error detection and handling. Table~\ref{tab:dataset_perturbations} summarises these perturbations and reports the uncertainty score for each run, with derivations in Appendix~\ref{sec:appA}. Following \cite{Oreski2017}, these scores help examine how dataset characteristics relate to workflow uncertainty.

Appendix~\ref{sec:appB} compares the uncertainty distributions of default and perturbed runs for the Qwen backend, which showed the strongest error detection and adaptability. Runs with detected issues may fall outside the default distribution, while the scenarios in Table~\ref{tab:workflow_scenarios} are shown as individual points. Large deviations, for example measured using conformal prediction, could serve as warning signals for further review.

\section{Conclusion}

This work presented a runtime uncertainty-monitoring framework for LLM-based multi-agent systems in actuarial risk prediction. The system uses specialised agents for data preparation, modelling, review, and explanation, coordinated through a central hub. Its main contribution is the use of calibrated token-level log-probability signals together with a Bayesian Network to model how uncertainty propagates across agent outputs and workflow stages. Rather than assessing each agent in isolation, the framework traces uncertainty through the full workflow, helping to identify weak points and cases where human review may be needed. This is particularly important in actuarial settings, where unreliable outputs may lead to incorrect risk assessment, unfair pricing, or regulatory concerns.

The results show that the MAS can reproduce baseline actuarial modelling performance while providing additional insight into workflow-level confidence. However, the effectiveness of the approach depends on the selected LLM backend, with differences observed in error detection, adaptability, and uncertainty behaviour. The proposed Bayesian Network should therefore be interpreted as a runtime uncertainty-propagation monitor, not as a proof of output correctness.

This study has several limitations. The Bayesian Network abstracts the workflow as a directed acyclic dependency model and does not explicitly represent execution sequence, task priority, repeated revisions, or feedback loops between agents. These aspects are handled operationally by the central hub, while the Bayesian Network uses the final accepted output of each stage as evidence. Future work will extend the framework to better distinguish intrinsic and extrinsic uncertainty \cite{Duan2025}, introduce symptom layers for identifying the causes of uncertainty \cite{LIU20151917,he2025}, and explore dynamic Bayesian Networks to capture temporal behaviour and iterative repair cycles.

\begin{credits}
\subsubsection{Data and Code Availability}
Regarding research reproducibility, the Python implementation, datasets, and evaluation notebooks supporting this paper are publicly available on GitHub: \url{https://github.com/bart-custers/actuarial_agents}

\subsubsection{\discintname}
The authors have no competing interests.
\end{credits}
%
%
%
\bibliographystyle{splncs04}
\bibliography{references}

@article{Owens2022,
   author = {Emer Owens and Barry Sheehan and Martin Mullins and Martin Cunneen and Juliane Ressel and German Castignani},
   doi = {10.3390/risks10120230},
   issn = {22279091},
   journal = {Risks},
   month = {12},
   pages = {230},
   title = {Explainable Artificial Intelligence (XAI) in Insurance},
   volume = {10},
   year = {2022}
}

@article{Wang2024,
   author = {Lei Wang and Chen Ma and Xueyang Feng and Zeyu Zhang and Hao Yang and Jingsen Zhang and Zhiyuan Chen and Jiakai Tang and Xu Chen and Yankai Lin and Wayne Xin Zhao and Zhewei Wei and Jirong Wen},
   doi = {10.1007/s11704-024-40231-1},
   issn = {20952236},
   issue = {6},
   journal = {Frontiers of Computer Science},
   month = {12},
   publisher = {Higher Education Press Limited Company},
   title = {A survey on large language model based autonomous agents},
   volume = {18},
   year = {2024}
}

@article{Sapkota2025,
   author = {Ranjan Sapkota and Konstantinos I. Roumeliotis and Manoj Karkee},
   doi = {10.48550/arXiv.2505.10468},
   journal = {arXiv 2505.10468},
   month = {5},
   title = {AI Agents vs. Agentic AI: A Conceptual Taxonomy, Applications and Challenges},
   year = {2025}
}

@inproceedings{Mohammadi2025,
   author = {Mahmoud Mohammadi and Yipeng Li and Jane Lo and Wendy Yip},
   city = {Toronto, Canada},
   doi = {10.1145/3711896.3736570},
   booktitle = {Proceedings of the 31st ACM SIGKDD Conference on Knowledge Discovery and Data Mining V.2},
   month = {8},
   pages = {6129-6139},
   publisher = {Association for Computing Machinery (ACM)},
   title = {Evaluation and Benchmarking of LLM Agents: A Survey},
   year = {2025}
}

@article{Yehudai2025,
   author = {Asaf Yehudai and Lilach Eden and Alan Li and Guy Uziel and Yilun Zhao and Roy Bar-Haim and Arman Cohan and Michal Shmueli-Scheuer},
   doi = {10.48550/arXiv.2503.16416},
   journal = {arXiv 2503.16416},
   month = {3},
   title = {Survey on Evaluation of LLM-based Agents},
   year = {2025}
}

@article{Kim2017,
   author = {Been Kim and Martin Wattenberg and Justin Gilmer and Carrie Cai and James Wexler and Fernanda Viegas and Rory Sayres},
   doi = {10.48550/arXiv.1711.11279},
   journal = {arXiv 1711.11279},
   month = {6},
   title = {Interpretability Beyond Feature Attribution: Quantitative Testing with Concept Activation Vectors (TCAV)},
   year = {2017}
}

@article{Chen2023,
   author = {Weize Chen and Yusheng Su and Jingwei Zuo and Cheng Yang and Chenfei Yuan and Chi-Min Chan and Heyang Yu and Yaxi Lu and Yi-Hsin Hung and Chen Qian and Yujia Qin and Xin Cong and Ruobing Xie and Zhiyuan Liu and Maosong Sun and Jie Zhou},
   doi = {10.48550/arXiv.2308.10848},
   journal = {arXiv 2308.10848},
   month = {10},
   title = {AgentVerse: Facilitating Multi-Agent Collaboration and Exploring Emergent Behaviors},
   year = {2023}
}

@article{Wu2023,
   author = {Qingyun Wu and Gagan Bansal and Jieyu Zhang and Yiran Wu and Beibin Li and Erkang Zhu and Li Jiang and Xiaoyun Zhang and Shaokun Zhang and Jiale Liu and Ahmed Hassan Awadallah and Ryen W White and Doug Burger and Chi Wang},
   doi = {10.48550/arXiv.2308.08155},
   journal = {arXiv 2308.08155},
   month = {10},
   title = {AutoGen: Enabling Next-Gen LLM Applications via Multi-Agent Conversation},
   year = {2023}
}

@article{Yang2024,
   author = {Yingxuan Yang and Qiuying Peng and Jun Wang and Ying Wen and Weinan Zhang},
   doi = {10.48550/arXiv.2411.14033},
   journal = {arXiv 2411.14033},
   month = {12},
   title = {LLM-based Multi-Agent Systems: Techniques and Business Perspectives},
   year = {2024}
}

@article{Tran2025,
   author = {Khanh-Tung Tran and Dung Dao and Minh-Duong Nguyen and Quoc-Viet Pham and Barry O'Sullivan and Hoang D. Nguyen},
   doi = {10.48550/arXiv.2501.06322},
   journal = {arXiv 2501.06322},
   month = {1},
   title = {Multi-Agent Collaboration Mechanisms: A Survey of LLMs},
   year = {2025}
}

@article{InternationalActuarialAssociation2024,
   author = {{International Actuarial Association}},
   title = {Artificial Intelligence Governance Framework - General Actuarial Practice},
   year = {2024},
   journal = {https://actuaries.org/paper/artificial-intelligence-governance-framework/ [Accessed 3/12/25]}
}

@article{Hatzesberger2025,
   author = {Simon Hatzesberger and Iris Nonneman},
   doi = {10.48550/arXiv.2506.18942},
   journal = {arXiv 2506.18942},
   month = {6},
   title = {Advanced Applications of Generative AI in Actuarial Science: Case Studies Beyond ChatGPT},
   year = {2025}
}

@article{Qwen,
   author = {{Qwen Team}},
   title = {Qwen2.5-7B-Instruct (model card)},
   journal = {https://huggingface.co/Qwen/Qwen2.5-7B-Instruct [Accessed 25/1/2026]},
   year = {2025}
}

@article{MetaAI2024,
   author = {{Meta AI}},
   month = {7},
   title = {Introducing Llama 3.1},
   journal = {https://ai.meta.com/blog/meta-llama-3-1/ [Accessed 25/1/2026]},
   year = {2024}
}

@inproceedings{Xu2025,
   author = {Mengyao Xu and Qiaoyin Gan and Zhenyu Zhu and Haojun Qin},
   city = {Trondheim, Norway},
   doi = {10.1145/3696630.3731433},
   isbn = {9798400712760},
   issn = {15397521},
   booktitle = {Proceedings of the ACM SIGSOFT Symposium on the Foundations of Software Engineering},
   month = {7},
   pages = {1242-1243},
   publisher = {Association for Computing Machinery},
   title = {Logprobs Know Uncertainty: Fighting LLM Hallucinations},
   year = {2025}
}

@article{Chauvin2025,
   author = {Timothée Chauvin and Erwan Le Merrer and François Taïani and Gilles Tredan},
   doi = {10.48550/arXiv.2512.03816},
   journal = {arXiv 2512.03816},
   month = {12},
   title = {Log Probability Tracking of LLM APIs},
   year = {2025}
}

@book{Pearl1988,
   author = {J. Pearl},
   doi = {10.1016/C2009-0-27609-4},
   publisher = {Morgan Kaufmann Publishers Inc., San Francisco, USA},
   title = {Probabilistic Reasoning in Intelligent Systems},
   year = {1988}
}

@article{Nafar2025,
   author = {Aliakbar Nafar and Kristen Brent Venable and Zijun Cui and Parisa Kordjamshidi},
   doi = {10.48550/arXiv.2505.15918},
   journal = {arXiv 2505.15918},
   month = {8},
   title = {Extracting Probabilistic Knowledge from Large Language Models for Bayesian Network Parameterization},
   year = {2025}
}

@article{Hu2024,
   author = {Zhengmian Hu and Tong Zheng and Heng Huang},
   doi = {10.48550/arXiv.2410.21716},
   journal = {arXiv 2410.21716},
   month = {10},
   title = {A Bayesian Approach to Harnessing the Power of LLMs in Authorship Attribution},
   year = {2024}
}

@article{Duan2025,
   author = {Jinhao Duan and James Diffenderfer and Sandeep Madireddy and Tianlong Chen and Bhavya Kailkhura and Kaidi Xu},
   doi = {10.48550/arXiv.2506.17419},
   journal = {arXiv 2506.17419},
   month = {6},
   title = {UProp: Investigating the Uncertainty Propagation of LLMs in Multi-Step Agentic Decision-Making},
   year = {2025}
}

@article{Oreski2017,
   author = {Dijana Oreski and Stjepan Oreski and Bozidar Klicek},
   doi = {10.1016/j.asoc.2016.12.023},
   issn = {15684946},
   journal = {Applied Soft Computing},
   month = {3},
   pages = {109-119},
   publisher = {Elsevier Ltd},
   title = {Effects of dataset characteristics on the performance of feature selection techniques},
   volume = {52},
   year = {2017}
}

@article{Malinin2020,
   author = {Andrey Malinin and Mark Gales},
   doi = {10.48550/arXiv.2002.07650},
   journal = {arXiv 2002.07650},
   month = {2},
   title = {Uncertainty Estimation in Autoregressive Structured Prediction},
   year = {2020}
}

@article{LIU20151917,
title = {An approach for developing diagnostic Bayesian network based on operation procedures},
journal = {Expert Systems with Applications},
volume = {42},
number = {4},
pages = {1917-1926},
year = {2015},
issn = {0957-4174},
doi = {https://doi.org/10.1016/j.eswa.2014.10.020},
url = {https://www.sciencedirect.com/science/article/pii/S0957417414006447},
author = {Zengkai Liu and Yonghong Liu and Baoping Cai and Chao Zheng},
}

@article{he2025,
      title={SentinelAgent: Graph-based Anomaly Detection in Multi-Agent Systems}, 
      author={Xu He and Di Wu and Yan Zhai and Kun Sun},
      year={2025},
      journal={arXiv 2505.24201},
      doi={10.48550/arXiv.2505.24201}, 
}

@article{donaldson2026bayesian,
  title={Bayesian Uncertainty Propagation for Agentic RAG Pipelines: A Proof-of-Concept Study on Multi-Hop Question Answering},
  author={Donaldson, Louis and Walker, Connor and Aslansefat, Koorosh and Papadopoulos, Yiannis},
  journal={arXiv preprint arXiv:2607.00972},
  year={2026}
}

\appendix
\section{Appendix: Dataset severity indicators} 
\label{sec:appA}

To quantify how strongly a dataset has changed between a baseline dataset \(D_b\) and a perturbed dataset \(D_p\), several indicators were computed. These indicators capture structural, statistical, and distributional changes between the datasets.

\subsection{Row Change}
The row change indicator measures the relative change in the number of observations between the baseline and perturbed dataset.

\begin{equation}
\text{RowChange} =
\frac{|n_p - n_b|}{n_b}
\end{equation}
where:
\begin{itemize}
\item \(n_b\) is the number of rows in the baseline dataset,
\item \(n_p\) is the number of rows in the perturbed dataset.
\end{itemize}

\subsection{Column Change}
The column change indicator measures the relative change in the number of features (columns).

\begin{equation}
\text{ColChange} =
\frac{|p_p - p_b|}{p_b}
\end{equation}
where:
\begin{itemize}
\item \(p_b\) is the number of columns in the baseline dataset,
\item \(p_p\) is the number of columns in the perturbed dataset.
\end{itemize}

\subsection{Missing values}
The missing value indicator measures how the total number of missing values changes between datasets relative to the number of cells. It is computed as:

\begin{equation}
\text{Missingness} =
\frac{|M(D_p) - M(D_b)|}{n_b p_b}
\end{equation}
where:
\begin{itemize}
\item \(M(D_p)\) is the number of missing values in the perturbed dataset,
\item \(M(D_b)\) is the number of missing values in the baseline dataset.
\end{itemize}

\subsection{Target Shift}
Target shift measures the standardized difference in the mean of the target variable between datasets.
Let \(Y\) denote the target variable (in this case the number of claims). 
Define the mean of the target value as \(\mu_b = \mathbb{E}{D_b}[Y]\) and \(\mu_p = \mathbb{E}{D_p}[Y]\).
The target shift is then defined as:

\begin{equation}
\text{TargetShift} =
\frac{|\mu_p - \mu_b|}{\sigma_b}
\end{equation}
where:
\begin{itemize}
\item \(mu_p\) is the mean target value in the perturbed dataset,
\item \(mu_b\) is the mean target value in the baseline dataset,
\item \(sigma_b\) is the standard deviation of the target value in the baseline dataset.
\end{itemize}

\subsection{Sparsity}
Sparsity measures the proportion of elements that are either zero or missing.
Let the number of zero values \(Z(D)\) and the number of missing values \(M(D)\) be defined by:
\[
Z(D) = \sum_{i=1}^{n}\sum_{j=1}^{p} \mathbf{1}(x_{ij} = 0), \qquad
M(D) = \sum_{i=1}^{n}\sum_{j=1}^{p} \mathbf{1}(x_{ij} = \text{NA})
\]
The sparsity of dataset \(D\) is defined as:
\begin{equation}
\text{Sparsity}(D) =
\frac{Z(D) + M(D)}{n p}
\end{equation}
where \(np\) is the total number of elements in the dataset.

\subsection{Population Shift}
Population shift is measured using a population stability index (PSI), which quantifies changes in the distribution of a variable between datasets.
First, the baseline variable is divided into \(K\) quantile-based bins.
Let $N^b_k$ and $N^p_k$ be the numbers of
observations from the base and perturbed datasets, respectively, that fall
into bin $I_k$. The corresponding proportions are
\[
p^b_k = \frac{N^b_k}{n_{\text{b}}}, \qquad
p^p_k  = \frac{N^p_k}{n_{\text{n}}}, \quad
k=1,\dots,K.
\]
The PSI is then defined as
\begin{equation}
\text{PSI} =
\sum_{k=1}^{K}
(p^p_k - p^b_k)\log\left(\frac{p^p_k}{p^b_k}\right)
\end{equation}
A higher PSI indicates stronger distributional drift.

\subsection{Correlation Shift}
Correlation shift is measured by how much the correlation structure between the numeric columns of two data sets changes. It computes the correlation matrix of all numeric variables in each dataset \(D\) and then returns the Euclidean (L2) norm of the difference between these two correlation matrices (i.e., a single number summarizing the overall shift in pairwise correlations):

\begin{equation}
\text{CorrShift} =
\left\|
R_p - R_b
\right\|
\end{equation}

where:
\begin{itemize}
\item \(R_b = \text{corr}(D_b)\) is the correlation matrix of the baseline dataset,
\item \(R_p = \text{corr}(D_p)\) is the correlation matrix of the perturbed dataset,
\item \(\left\| . \right\|\) is the default matrix 2-norm.
\end{itemize}

\section{Appendix: Uncertainty distribution} 
\label{sec:appB}

\begin{figure}[htbp!] 
\centering    
\includegraphics[width=0.95\textwidth]{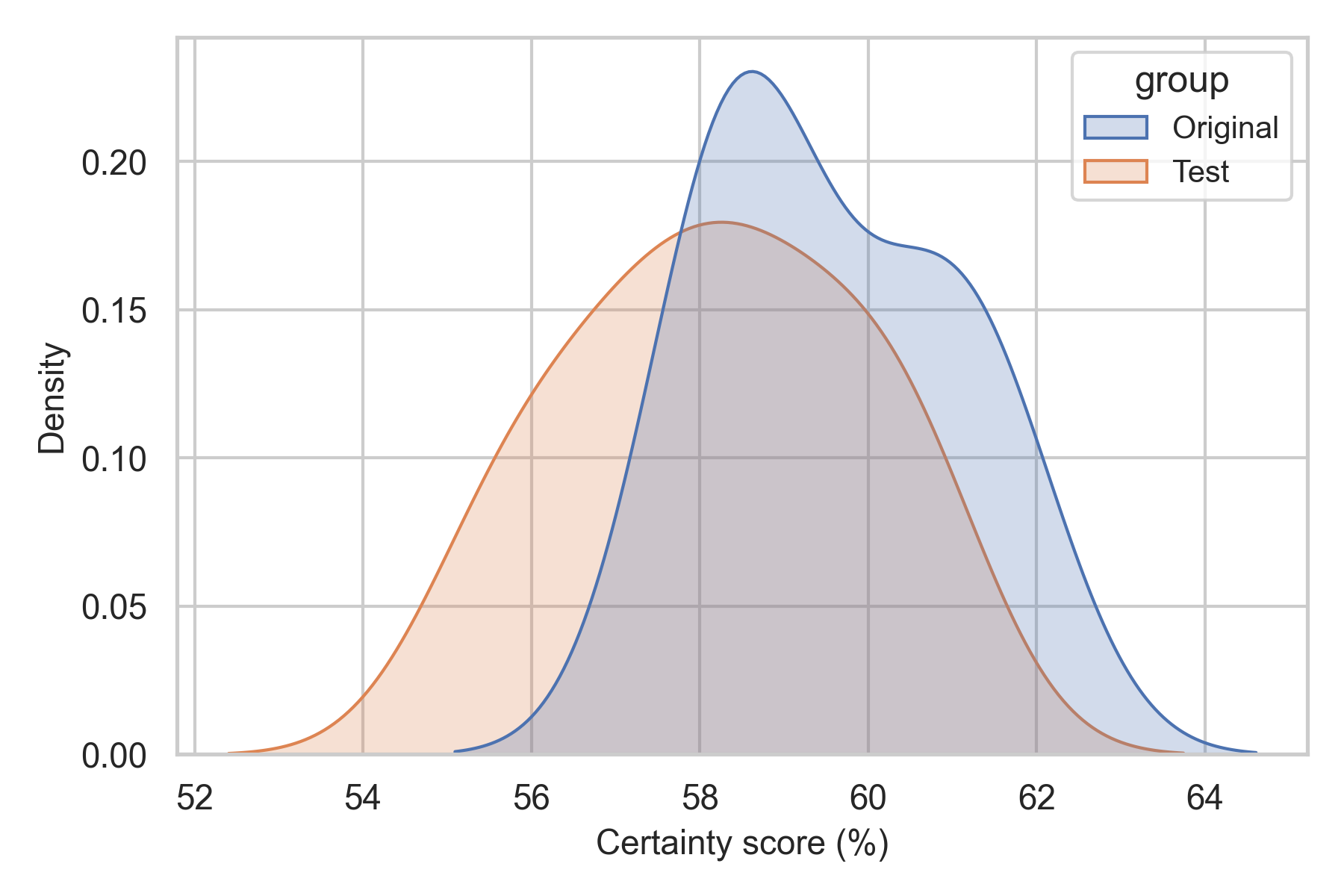}
\caption{Uncertainty distribution, compared between original (default) and test datasets, for workflow runs with Qwen backend. The four individual points represent the uncertainty outcomes from the test scenarios in Table \ref{tab:workflow_scenarios}.}
\label{fig:5.1}
\end{figure}

\end{document}